\documentclass[journal,onecolumn,11pt]{IEEEtran}

\usepackage{lineno,hyperref}
\usepackage{bm}
\usepackage{mathrsfs}
\usepackage{subfigure}
\usepackage{amssymb}
\usepackage{amsmath}
\usepackage{graphicx}
\usepackage{multirow}
\usepackage{textcomp,booktabs}
\usepackage[usenames,dvipsnames]{color}
\usepackage{colortbl}
\definecolor{mygray}{gray}{.9}
\usepackage{epstopdf}
\modulolinenumbers[5] 

\ifCLASSINFOpdf

\else

\fi

\begin{document}

\title{Conjugate Gradient Adaptive Learning with Tukey's Biweight M-Estimate}

\author{Lu Lu, Yi Yu, Rodrigo C. de Lamare,~\IEEEmembership{Senior Member, IEEE,} and Xiaomin Yang

    \thanks{L. Lu and X. Yang are with the College of Electronics and Information Engineering, Sichuan University, Chengdu 610065, China. (e-mail: lulu19900303@126.com, arielyang@scu.edu.cn).}
    \thanks{Y. Yu is with the School of Information Engineering, Robot Technology Used for Special Environment Key Laboratory of Sichuan Province, Southwest University of Science and Technology, Mianyang 621010, China. (e-mail: yuyi\_xyuan@163.com).}
    \thanks{R. C. de Lamare is with CETUC, PUC-Rio, Rio de Janeiro 22451-900, Brazil. (e-mail:
    delamare@cetuc.puc-rio.br).}}


\maketitle

\begin{abstract}
We propose a novel M-estimate conjugate gradient (CG) algorithm,
termed Tukey's biweight M-estimate CG (TbMCG), for system
identification in impulsive noise environments. In particular, the
TbMCG algorithm can achieve a faster convergence while retaining a
reduced computational complexity as compared to the recursive
least-squares (RLS) algorithm. Specifically, the Tukey's biweight
M-estimate incorporates a constraint into the CG filter to tackle
impulsive noise environments. Moreover, the convergence behavior of
the TbMCG algorithm is analyzed. Simulation results confirm the
excellent performance of the proposed TbMCG algorithm for system
identification and active noise control applications.
\end{abstract}


\section{Introduction}

\IEEEPARstart{T}{he} $\alpha$-stable noise is an effective approach
for modeling phenomena such as impulsive noise. In these scenarios,
the least mean square (LMS) type algorithms, which are based on the
mean square error (MSE) criterion, may fail to work
\cite{shao1993signal}. To address this problem, a number of
algorithms were developed
\cite{chambers1997robust,papoulis2004normalized,al2016robust,aydin1999robust,zayyani2014continuous,lu2018distributed,zayyani2020robust,zayyani2021robust,yu2021robust},
\cite{jidf,spa,intadap,mbdf,jio,jiols,jiomimo,sjidf,ccmmwf,tds,mfdf,l1stap,mberdf,jio_lcmv,locsme,smtvb,ccmrls,dce,itic,jiostap,aifir,ccmmimo,vsscmv,bfidd,mbsic,wlmwf,bbprec,okspme,rdrcb,smce,armo,wljio,saap,vfap,saalt,mcg,sintprec,stmfdf,1bitidd,jpais,did,rrmber,memd,jiodf,baplnc,als,vssccm,doaalrd,jidfecho,dcg,rccm,ccmavf,mberrr,damdc,smjio,saabf,arh,lsomp,jrpaalt,smccm,vssccm2,vffccm,sor,aaidd,lrcc,kaesprit,lcdcd,smbeam,ccmjio,wlccm,dlmme,listmtc,smcg,mfsic,cqabd,rmmse,rsthp,dmsmtc,dynovs,dqalms,detmtc,1bitce,mwc,dlmm,rsbd,rdcoprime,rdlms,lbal,wlbd,rrser}.
In \cite{arikan1995adaptive}, the normalized least mean $p$th power
(NLMP) algorithm was proposed, which is derived from the fractional
lower order moments (FLOM) of the error signal. The M-estimate based
algorithms (Huber's and Hample's) are effective methods to suppress
the effect of impulsive noise on the filter weights
\cite{petrus1999robust,zou2000least,zhou2010new,chan2011new}. Such
algorithms are essentially hybrid techniques based on $L_1$ and
$L_2$-norms and several variants have been presented
\cite{sun2015a,wu2013an,zhao2021robust}. However, these algorithms
can only deal with impulsive noise and cannot achieve improved
performance in Gaussian scenarios. For performance improvement, the
maximum correntropy criterion (MCC)-based algorithms were developed
\cite{chen2014steady,chen2012maximum,qian2018convergence}. Such MCC
method is also an M-estimator in essence \cite{liu2006error}. It has
revealed its effectiveness for non-Gaussian signal processing and
has been successfully applied in many applications
\cite{shi2020separable,kurian2017robust,zhao2022sb}. Note that the
above-mentioned algorithms rely on higher-order moments and
stochastic gradient (SG) methods to obtain good performance in
impulsive noise.

On the other hand, the recursive least-squares (RLS) algorithm offers considerable enhancement in convergence speed and tracking capability \cite{he2019efficient,bhotto2012new,aslam2019robust}. Regrettably, its heavy computational load prohibits its practical use. The conjugate gradient (CG) method, which can be viewed as an alternative to RLS, is an efficient implementation for adaptive filters \cite{lee2020sparse,Teoh1998Active}. By making use of the orthogonal search direction, such method can achieve faster convergence than the SG method \cite{xu2016distributed}. Recently, the CG method and its variants have received increased attention in kernel adaptive filters \cite{zhang2018kernel,xiong2020kernel} and beamforming \cite{wang2010constrained,wang2012set}. However, there is scarce literature focused on using the CG-type algorithm for combating impulsive noise.

Motivated by the above considerations, in this paper, we propose a novel CG algorithm for combating impulsive noise. The proposed algorithm embeds the CG method inside of the Tukey's biweight M-estimator \cite{beaton1974fitting,tyler2008robust}, resulting in the TbMCG algorithm. Compared to the above-mentioned algorithms, the proposed TbMCG algorithm is able to mitigate various noises, such as $\alpha$-stable noise, Gaussian noise, and mixed-noise. Moreover, the TbMCG algorithm can be easily extended to active noise control (ANC) system, generating a filtered-x TbMCG (FxTbMCG) algorithm. The FxTbMCG algorithm can also achieve improved performance as compared to the existing ANC algorithms.

\section{Problem formulation}
\label{sec2}
\begin{figure}[!htbp]
    \centering
    \includegraphics[scale=0.95]{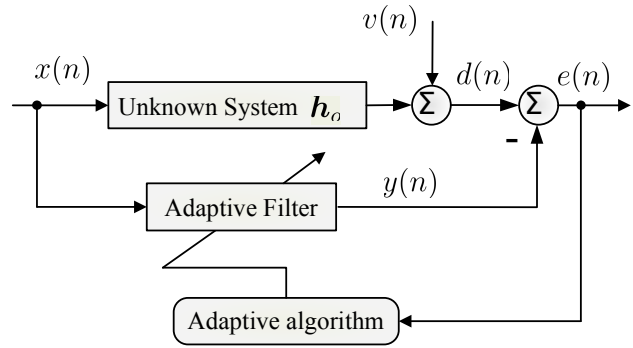}
    \caption{Diagram of the system identification problem.}
    \label{Fig01}
\end{figure}

Fig. \ref{Fig01} shows the system identification problem, where $x(n)$ denotes the input signal, $e(n)$ denotes the error signal, $\bm h_o$ denotes the weight vector of the unknown system, $v(n)$ denotes the impulsive noise, and $n$ denotes the time instant. The desired signal $d(n)$ is given by
\begin{equation}
d(n) = {\bm h}^{\mathrm T}_o{\bm x}(n) + v(n)
\label{001}
\end{equation}
where superscript $(\cdot)^{\mathrm T}$ stands for the transpose, $\bm x(n)=[x(n),x(n-1),\dots,x(n-L+1)]^{\mathrm T}$ represents the input vector, and $L$ is the filter length. The error signal $e(n)$ is defined as
\begin{equation}
e(n) = d(n) - y(n) = d(n) - {\bm h}^{\mathrm T}(n){\bm x}(n)
\label{002}
\end{equation}
where ${\bm h}(n)$ is the weight vector of the adaptive filter and $y(n)={\bm h}^{\mathrm T}(n){\bm x}(n)$. In this paper, the impulsive noise is generated by standard symmetric $\alpha$-stable ($S\alpha S$) distribution, whose characteristic function is expressed as $\vartheta(t) = \mathrm{exp}\left\{-|t|^\alpha\right\}$ \cite{lu2018distributed,shao1993signal}, where $0<\alpha\le2$ is the \emph{characteristic exponent}. A small value of $\alpha$ implies a strong impulsive process. Particularly, the $\alpha$-stable noise degenerates into Gaussian noise for $\alpha=2$.

\section{Proposed TbMCG algorithm}
\label{sec3}
In this section, we derive the proposed TbMCG algorithm and detail its computational cost.
\subsection{Derivation of TbMCG algorithm}

Considering the data set $\{\bm x(n), d(n)\}$, the objective function of Tukey's biweight M-estimate with variable $e$ is expressed as \cite{beaton1974fitting,tyler2008robust}
\begin{equation}
{\mathcal J}(e) = \left\{ \begin{array}{l}
\frac{e^2}{c^2} - \frac{e^4}{c^4} + \frac{e^6}{3c^6} \;\;\;\;\;\;\;\;{\rm {if}}\;|e| \leq c \\
\frac{1}{3} \;\;\;\;\;\;\;\;\;\;\;\;\;\;\;\;\;\;\;\;\;\;\;\;\;\;\;{\rm {if}}\;|e| > c \\
\end{array} \right.
\label{008}
\end{equation}
where $c$ is the positive constant. Recalling other M-estimate methods, Huber and Hample estimators, the non-negative threshold provides the control capability of constraining large outliers \cite{yu2019m}. The Huber estimator is based on a convex loss function, so it has no local minima. However, it does not emphasize large errors as compared to the MSE criterion $e^2$. The Hample's method, whose cost function is non-convex and can tackle extreme outliers better than the Huber one. The Tukey's biweight (or bisquare) M-estimator shares a similar property of Hample's M-estimator, whose objective function is also non-convex and can effectively suppress outliers. On other hand, it retains the merits of Huber's method. It has the simplicity of an expression depending on just one tuning parameter $c$, similar to the Huber's loss function \cite{tyler2008robust}.
\begin{figure}[!htbp]
    \centering
    \includegraphics[scale=0.5]{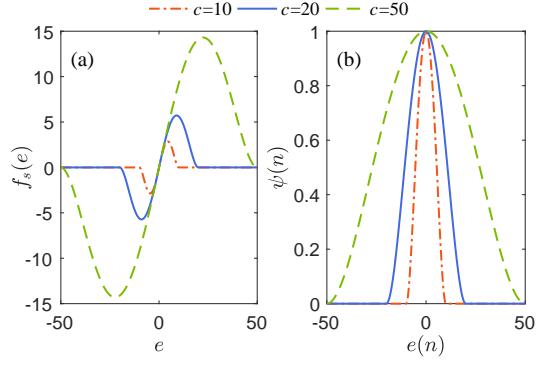}
    \caption{Evolution curves of the score function and the weighting factor.}
    \label{Fig02}
\end{figure}

The score function $f_s(e) = \partial{\mathcal J(e)}/\partial e$ is defined as
\begin{equation}
f_s(e) = \left\{ \begin{array}{l}
e\left[1-(\frac{e}{c})^2\right]^2 \;\;\;\;\;\;\;\;\;\;{\rm {if}}\;|e| \leq c \\
0 \;\;\;\;\;\;\;\;\;\;\;\;\;\;\;\;\;\;\;\;\;\;\;\;\;\;\;\;{\rm {if}}\;|e| > c. \\
\end{array} \right.
\label{009}
\end{equation}
To show the property of the score function, Fig. \ref{Fig02} (a) plots the evolution curve of $f_s(e)$. One can observe from this figure that the score function owns a `redescend' feature, that is, when $|e|$ is large enough, $f_s(e)$ tends to zero. Such a phenomenon implies Tukey's biweight M-estimate is insensitive to the outliers. Note that (\ref{009}) can save computational load $f_s(e)=0$ when the large outliers occurs. Moreover, the score function is linear at $e=0$ in accordance with Winsor's principle that all distributions are normal in the middle \cite{hardin2000multivariate}.


Taking the gradient of the score function w.r.t. $\bm h(n)$, yields
\begin{equation}
\frac{1}{K} \sum\limits_{n=1}^K \psi(n)\bm x(n)\bm x^{\mathrm T}(n) \bm h(n) = \frac{1}{K} \sum\limits_{n=1}^K \psi(n)d(n)\bm x^{\mathrm T}(n)
\label{010}
\end{equation}
where $K$ denotes the sampling points and $\psi(n)$ is the weighting factor, which is expressed as
\begin{equation}
\psi(n) = \left\{ \begin{array}{l}
\left[1-\left(\frac{e(n)}{c}\right)^2\right]^2 \;\;\;\;\;\;\;\;{\rm {if}}\;|e(n)| \leq c \\
0 \;\;\;\;\;\;\;\;\;\;\;\;\;\;\;\;\;\;\;\;\;\;\;\;\;\;\;\;\;\;{\rm {otherwise}}. \\
\end{array} \right.
\label{011}
\end{equation}
Fig. \ref{Fig02} (b) depicts the curve of $\psi(n)$. As can be seen, the amplitude of $\psi(n)$ is restricted from the range of $[0,1]$, contributing to a good performance for attenuating outliers. For simplicity, define $\bm R = \frac{1}{K} \sum\nolimits_{n=1}^K \psi(n)\bm x(n)\bm x^{\mathrm T}(n)$ as the correlation matrix of the input signal and $\bm \theta = \frac{1}{K} \sum\nolimits_{n=1}^K \psi(n) d(n)\bm x^{\mathrm T}(n)$ as the cross-correlation vector between the desired signal and the input signal. Akin to the RLS and Newton method, the CG method can iteratively solve for the optimal solution $\bm R^{-1}(n)\bm \theta(n)$. The expression of $\bm R(n)$ and $\bm \theta(n)$ for the TbMCG algorithm is given by
\begin{equation}
\bm R(n) = \lambda \bm R(n-1) + \psi(n) \bm x(n)\bm x^{\mathrm T}(n)
\label{013}
\end{equation}
and
\begin{equation}
\bm \theta(n) = \lambda \bm \theta(n-1) + \psi(n) d(n)\bm x^{\mathrm T}(n)
\label{014}
\end{equation}
where $0 \ll \lambda < 1$ is  the the forgetting factor. When we set $\psi(n)=1$, the TbMCG algorithm reduces to the standard CG algorithm \cite{chang2000analysis}.

The TbMCG algorithm computes $\bm h(n)$ via $\bm h(n-1)$ with the step size $\delta(n)$ at previous iteration $n-1$. By exploiting the orthogonality of the residuals and these previous search directions, the search linearity is independent of the previous directions. For the next solution, a new search direction,  a residual, and a step size will be calculated. The residual vector can be expressed as
\begin{equation}
\begin{aligned}
\bm g(n) = \bm \theta(n) - \bm R(n)\bm h(n)
=& \lambda \bm g(n-1) - \delta(n) \bm R(n) \bm p(n-1) \\
&+ \psi(n)\bm x(n) e(n).
\end{aligned}
\label{017}
\end{equation}
Left multiplying $\bm p^{\mathrm T}(n-1)$ on both sides of (\ref{017}) and taking the expectation operation on the both sides, we have
\begin{equation}
\begin{aligned}
{\mathrm E}\left\{ \bm p^{\mathrm T}(n-1)\bm g(n) \right\} \approx&\; \lambda{\mathrm E}\left\{ \bm p^{\mathrm T}(n-1) \bm g(n) \right\}  - {\mathrm E}\left\{ \delta(n) \right\}\\
&\times {\mathrm E}\left\{ \bm p^{\mathrm T}(n-1) \bm R(n) \bm p(n-1) \right\}  \\
&+ {\mathrm E}\left\{\psi(n)\right\} {\mathrm E}\left\{ \bm p^{\mathrm T}(n-1) \right\}\\
&\times {\mathrm E}\left\{ \bm x(n) \left[d(n)-\bm h^{\mathrm T}(n)\bm x(n)\right] \right\}
\end{aligned}
\label{019}
\end{equation}
where ${\mathrm E}\{\cdot\}$ denotes the expectation. Using the approximation results in \cite{xu2016distributed}, the step size $\delta(n)$ is defined by
\begin{equation}
\begin{aligned}
\delta(n) \triangleq \eta\frac{\bm p^{\mathrm T}(n-1)\bm g(n-1) }{ \bm p^{\mathrm T}(n-1) \bm R(n) \bm p(n-1) }
\end{aligned}
\label{020}
\end{equation}
where $\eta>0$ is the constant parameter. Thus, the search direction vector $\bm p(n)$ of the TbMCG algorithm is given by
\begin{equation}
\begin{aligned}
\bm p(n) = \bm g(n) + \beta(n)\bm p(n-1)
\end{aligned}
\label{021}
\end{equation}
where $\beta(n)$ is the coefficient parameter. For the TbMCG algorithm, the Polak \& Ribi{\`e}re (PR) approach is used \cite{peng2017kernel}
\begin{equation}
\begin{aligned}
\beta(n) = \frac{\left[\bm g(n) - \bm g(n-1)\right]^{\mathrm T}\bm g(n)}{\bm g^{\mathrm T}(n-1)\bm g(n-1) }.
\end{aligned}
\label{022}
\end{equation}
At last, the update equation of the weight vector for the TbMCG algorithm is
\begin{equation}
\begin{aligned}
\bm h(n) = \bm h(n-1) + \delta(n)\bm p(n).
\end{aligned}
\label{023}
\end{equation}


\begin{table}[htb]
    \small
    \centering
    \scriptsize
    \caption{Summary of the computational complexity.}
    \label{Table02}
    \doublerulesep=0.5pt
    \begin{tabular}{c|c|c|c}
        \hline
        \hline
        \textbf {Algorithms} &\textbf {$+/-$} &\textbf {$\times/\div$} &\textbf {Special instructions} \\ \hline
        \textbf{LMM}  &$2L+4$     &$2L+N_w+7$      &0 \\ \hline
        \textbf{RLS}  &$3L^2$     &$4L^2+3L+1$      &0 \\ \hline
        \textbf{NMCC}  &$2L-1$   &$2L+2$       &\begin{tabular}[c]{@{}l@{}}1 exponential operation\end{tabular} \\ \hline
        \textbf{CG}  &$2L^2+9L-4$     &$3L^2+10L+2$      &0 \\ \hline
        \textbf{TbMCG} &$2L^2+9L-3$ &$3L^2+12L+4$  &1 comparison\\ \hline
        \hline \hline
    \end{tabular}
\end{table}

The computational complexity in terms of the number of operations in each iteration is analyzed in this subsection. Table \ref{Table02} compares the computational costs of the least mean M-estimate (LMM) \cite{zou2000least}, RLS \cite{sayed2011adaptive}, normalized MCC (NMCC) \cite{kurian2017robust}\footnote{The NMCC algorithm in \cite{kurian2017robust} is applied to ANC. Such an algorithm can be easily derived for system identification.}, CG \cite{chang2000analysis} and the proposed algorithms, where $N_w$ denotes the window length of the estimation window. Considering the conventional RLS algorithm, it is obvious that the CG algorithm has a reduced computational cost. It can be seen that the dominant terms of the complexity of the CG and TbMCG algorithms are the same or similar. To compute the weighting factor $\psi(n)$, the TbMCG algorithm requires extra 1 addition, $2L+2$ multiplications, and 1 comparison. It should be emphasized that the computational complexity of the TbMCG algorithm may be lower than the CG algorithm, since $\psi(n)=0$ when the large outliers happen and the second term of (\ref{013}) and the third term of (\ref{019}) can be omitted.

\section{Convergence analysis}
\label{sec4}
In the TbMCG algorithm, the subspace spanned by the direction vectors is equal to the subspace spanned by the residuals. Furthermore, the residual $\bm g(n)$ can be regarded as a linear combination of the previous residuals and $\bm R\bm p(n)$. Revisiting $\bm p(n+1) \in \mathcal H(n)$, which indicates that each new $\mathcal H(n+1)$ is formed from the previous $\mathcal H(n)$ and $\bm R\mathcal H(n)$, where $\mathcal H(n)$ is the Krylov subspace spanning by $\{\bm p (n-1)\}$
\begin{equation}
\begin{aligned}
\mathcal H(n) \triangleq&\; {\mathrm {span}} \left\{ \bm p(1),\bm R\bm p(n),\ldots,\bm R^{n-1}\bm p(1) \right\}\\
=&\; {\mathrm {span}} \left\{ \bm g(0),\bm R\bm g(0),\ldots,\bm R^{n-1}\bm g(0) \right\}.
\end{aligned}
\label{024}
\end{equation}
From (\ref{017}), $\bm g(n)$ can be rewritten as $\bm g(n) = \bm R\bm h_o - \bm R\bm h(n)
=\bm R \widetilde {\bm h}(n)$, where $\bm h_o$ denotes the optimal solution of the weight vector and $\widetilde {\bm h}(n) \triangleq \bm h_o - \bm h(n)$ denotes the weight error vector. Then, (\ref{024}) can be rewritten by $\mathcal H(n) = {\mathrm {span}} \left\{ \bm R \widetilde {\bm h}(0),\bm R^2 \widetilde {\bm h}(0),\ldots,\bm R^n \widetilde {\bm h}(0) \right\}$. Substituting $\delta(n)$ and $\bm p(n+1)$ into $\bm h(n+1)$, we have
\begin{equation}
\begin{aligned}
\bm h(n+1) =&\; \bm h(0) + [\mathcal F_{\bm g_n}\bm g(n) + \mathcal F_{\bm g_{n-1}}\bm g(n-1)+\ldots\\
&+\mathcal F_{\bm g_1}\bm g(1) + \mathcal F_{\bm p_1}\bm p(1)]
\end{aligned}
\label{028}
\end{equation}
where $\mathcal F_{\bm g_n} = \delta(n+1)$, $\mathcal F_{\bm g_{n-1}} = \delta(n) + \delta(n+1)\beta(n)$, $\mathcal F_{\bm g_1} = \delta(2) + \delta(3)\beta(2)+\delta(4)\beta(3)\beta(2)+\ldots+\delta(n+1)\beta(n)\beta(n-1)\ldots\beta(2)$, $\mathcal F_{\bm p_1} = \delta(1) + \delta(2)\beta(1)+\delta(3)\beta(2)\beta(1)+\delta(n+1)\beta(n)\beta(n-1)\ldots\beta(1)$. Subtracting equation (\ref{028}) from $\bm h_o$, we have $\widetilde {\bm h}(n+1) = \widetilde {\bm h}(n) + \sum\limits_{j=1}^{n}\Psi(j)\bm R^j\widetilde {\bm h}(0)$, where $\Psi(j)$ is the linear combination of $\delta(l)$ and $\beta(k)$. It is shown that the Krylov subspace has the following property \cite{wang2010constrained}:
\begin{equation}
\begin{aligned}
\widetilde {\bm h}(n) = \left({\bf I} + \sum\limits_{j=1}^{n-1}\Psi(j)\bm R^j \right)\widetilde {\bm h}(0)
\end{aligned}
\label{031}
\end{equation}
where $\bf I$ stands for an identity matrix. Thus, (\ref{031}) is expressed in the form of a polynomial ${\mathcal L}_n(\bm R)$, as below
\begin{equation}
\begin{aligned}
\widetilde {\bm h}(n+1) = {\mathcal L}_n(\bm R)\widetilde {\bm h}(0)
\end{aligned}
\label{032}
\end{equation}
where we let ${\mathcal L}_0(\bm R)=\bf I$ due to the fact that the TbMCG algorithm cannot reach steady-state at the initial convergence. The vector $\widetilde {\bm h}(0)=\sum\limits_{j} \varrho_j \bm \varphi_j$, where $\varrho_j$ are the scalars not all zero and $j$ is the index number which relates to the number of the eigenvectors $\bm \varphi_j$. Considering ${\mathcal L}_n(\bm R)\bm \varphi_j = {\mathcal L}_n(\tau_j)\bm \varphi_j$, in which $\tau_j$ denotes the eigenvalue to the eigenvectors $\bm \varphi_j$, (\ref{032}) can be rewritten as $\widetilde {\bm h}(n+1) = \sum\limits_{j} \varrho_j{\mathcal L}_n(\tau_j)\bm \varphi_j$ and $|| \widetilde {\bm h}(n+1) ||_{\bm R}^2 = \sum\limits_{j} \varrho^2_j{\mathcal L}^2_n(\tau_j)\tau_j$. The essence of the TbMCG algorithm is to find the minimum value of $\mathcal L(\tau_j)$ during adaptation. In this regard, we have
\begin{equation}
\begin{aligned}
\left\| \widetilde {\bm h}(n+1) \right\|_{\bm R}^2 \leq&\; \mathop {\min}\limits_{\mathcal L_n} \mathop {\max}\limits_{\tau \in \Xi(\bm R)}{\mathcal L}^2_n(\tau)\sum\limits_{j} \varrho^2_j\tau_j\\
=&\; \mathop {\min}\limits_{\mathcal L_n} \mathop {\max}\limits_{\tau \in \Xi(\bm R)}{\mathcal L}^2_n(\tau) \left\| \widetilde {\bm h}(0) \right\|_{\bm R}^2
\end{aligned}
\label{035}
\end{equation}
where $\Xi(\bm R)$ represents the set of eigenvalues of $\bm R$. By using the Chebyshev polynomials \cite{watkins2004fundamentals}, we arrive at
\begin{equation}
\begin{aligned}
\left\| \widetilde {\bm h}(n) \right\|_{\bm R}^2 \leq&\; 2\left(\frac{\sqrt{\zeta}-1}{\sqrt{\zeta}+1}\right)^n \left\|\widetilde {\bm h}(0) \right\|_{\bm R}^2
\end{aligned}
\label{036}
\end{equation}
where $\zeta$ is the condition number. Owing to the fact that $\zeta$ is close to 1 for the optimal solution, and therefore $\left(\frac{\sqrt{\zeta}-1}{\sqrt{\zeta}+1}\right)^n$ is close to zero after iteration $n$. Hence, $\left\| \widetilde {\bm h}(n) \right\|_{\bm R}^2$ can converge to zero at iteration $n$.

\section{Simulation results}
\label{sec5}
In this section, we assess the proposed TbMCG algorithm in system identification and active noise control applications.
\subsection{System identification}
In this example, we compare the performance of the TbMCG algorithm with the LMM \cite{zou2000least}, NMCC \cite{kurian2017robust}, RLS \cite{sayed2011adaptive}, and CG \cite{chang2000analysis} algorithms in the context of system identification. The unknown system was a ten-tap filter generated randomly. The input is an AR(1) signal with a pole at 0.5. The normalized mean-square deviation (NMSD) ${\rm {NMSD}} = 20\log\left\{ {\left\| \bm h(n) - \bm h_o \right\|} / {\bm h_o} \right\}$ is used to evaluate the performance. Simulation results are obtained by averaging the curves over 100 independent runs.

\begin{figure}[!htbp]
    \centering
    \includegraphics[scale=0.5]{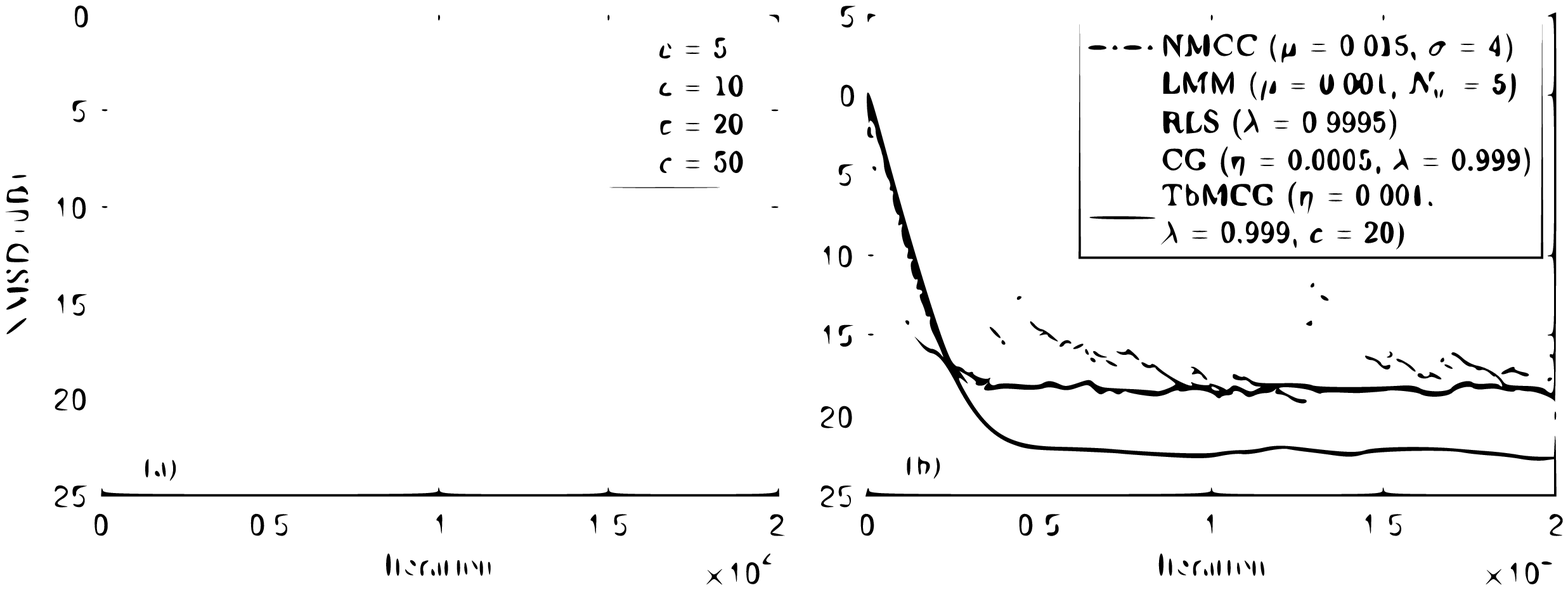}
    \caption{NMSDs of the algorithms under the colored input in $\alpha=1.8$. (a) TbMCG algorithm versus different $c$ ($\lambda=0.999$, $\eta=0.001$); (b) Comparison study.}
    \label{Fig03}
\end{figure}

\begin{figure}[!htbp]
    \centering
    \includegraphics[scale=0.5]{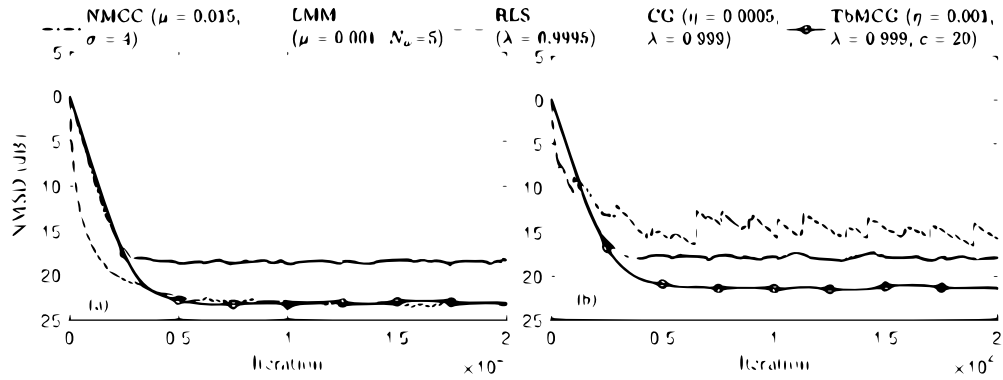}
    \caption{NMSDs of the algorithms under the colored input. (a) Gaussian noise environment ($\alpha=2$); (b) Mixed-noise (white Gaussian noise with signal-to-noise ratio (SNR)$=5$dB and impulsive noise with $\alpha=1.7$.}
    \label{Fig04}
\end{figure}

The effect of $c$ on the performance of the algorithm is shown in Fig. \ref{Fig03} (a) and a comparison study is conducted in Fig. \ref{Fig03} (b). As can be seen, $c$ behaves similar to the step-size of the LMS algorithm. A small value of $c$ can obtain small steady-state error whereas a large value of $c$ can achieve fast convergence. By considering the convergence rate and steady-state error, we select $c=20$ in this example. Moreover, it can be observed that the TbMCG algorithm has smaller NMSD for $\alpha=1.8$. We further assess the performance of those algorithms in Gaussian and mixed-noise environments, as illustrated in Fig. \ref{Fig04}, where the LMM algorithm has high steady-state misadjustment in the Gaussian scenario and the TbMCG algorithm achieves improved performance in both
cases.


\subsection{Active noise control application}
The TbMCG algorithm can be applied to the ANC problem, leading to the FxTbMCG algorithm. The RFxLMS \cite{george2012a}, FxRLS \cite{sayed2011adaptive} \footnote{The FxRLS algorithm can be easily obtained by extending the RLS algorithm to the ANC system. The RFxLMS algorithm can be obtained from the RFsLMS algorithm by removing the nonlinear expansion.}, FxlogLMS \cite{wu2011an}, and FxCG algorithms \cite{Teoh1998Active} are employed as benchmarks. Both primary and secondary paths are modeled by the finite impulse response (FIR) filter, obtained from \cite{kuo1996active}. To measure the reduction performance, the averaged noise reduction (ANR) is adopted \cite{wu2011an}. We set $L=128$ for all algorithms. Since the system model of ANC is totally different from system identification, the parameter setting in this example is different from Figs. \ref{Fig03} and \ref{Fig04}.

\begin{figure}[!htbp]
    \centering
    \includegraphics[scale=0.5]{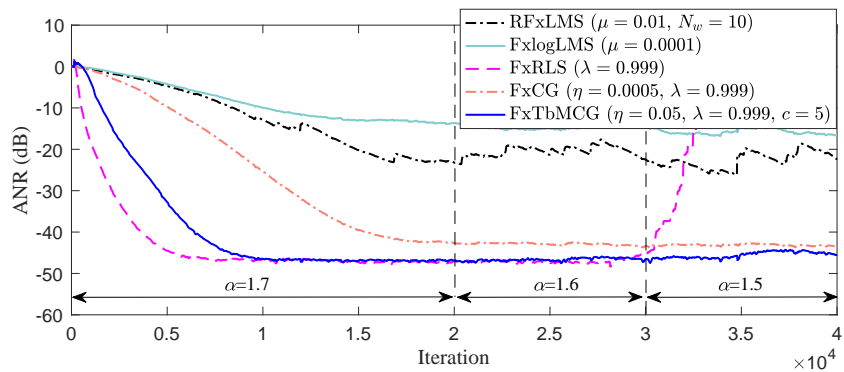}
    \caption{ANRs of five algorithms when the noise source is a time-varying $\alpha$-stable noise.}
    \label{Fig05}
\end{figure}
\begin{figure}[!htbp]
    \centering
    \includegraphics[scale=0.5]{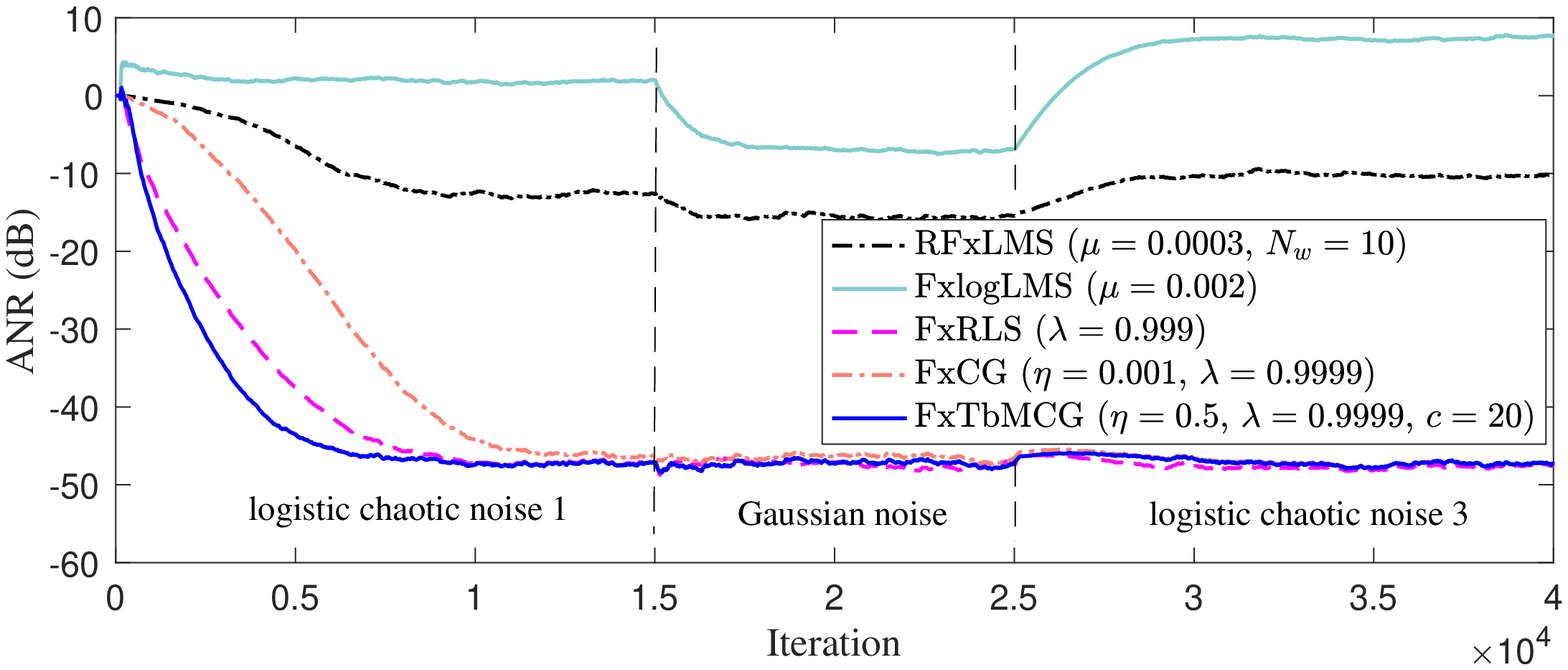}
    \caption{ANRs of five algorithms when the noise source is a time-varying noise.}
    \label{Fig06}
\end{figure}

First, the effect of time-varying noise sources in the attenuated performance is investigated in Figs. \ref{Fig05} and \ref{Fig06}. For the highly impulsive noise, $c$ is selected to be a small value. For the noise close to a Gaussian distribution, $c$ is selected to be a large value. Fig. \ref{Fig05} depicts the ANRs of five algorithms when the noise source is a time-varying $\alpha$-stable noise. It can be observed that the FxRLS algorithm diverges when $\alpha=1.5$. The FxTbMCG algorithm has a fast convergence after the abrupt changes at $n=0$, $n=20,000$ and $n=30,000$, achieving a small noise residual. Then, we focus on the case that the noise source is changed at iterations 15,000 and 25,000 and the algorithms run for 40,000 iterations. The used logistic chaotic noise 1 and 3 are generated by \cite{behera2014nonlinear}. As indicated in Fig. \ref{Fig06}, the RFxLMS algorithm cannot track the logistic chaotic noise 3 at iteration 25,000, and the FxTbMCG algorithm achieves good tracking performance and noise residual, followed by the FxRLS algorithm.

\section{Conclusions}
\label{sec6}
We have proposed the TbMCG algorithm by integrating the CG method with Tukey's biweight M-estimator for performance improvement in impulsive noise environments. Benefiting from the merits of the CG method and Tukey's biweight M-estimator, the proposed TbMCG algorithm can effectively restrict the outliers and has moderate computational complexity. The convergence of the TbMCG algorithm has been also studied. Simulation results in the context of system identification and active noise control have demonstrated that TbMCG yields smaller steady-state error and noise residual in comparison to state-of-the-art algorithms.

We also note that for Tukey's biweight M-estimator, $c$ is selected manually. The adaptation of $c$ is warranted for practical applications. The optimum parameter of $c$ can be obtained by the application of the methods in \cite{song2012normalized,chen2016kernel,huang2017adaptive}. A dedicated study will be conducted in future work.

\ifCLASSOPTIONcaptionsoff
\newpage
\fi

\footnotesize
\bibliographystyle{IEEEtran}
\bibliography{IEEEabrv,mybibfile}

\end{document}